\pdfoutput=1
\documentclass[11pt]{article}

\usepackage{coling}

\usepackage{times}
\usepackage{latexsym}

\usepackage[T1]{fontenc}
\usepackage[utf8]{inputenc}

\usepackage{microtype}

\usepackage{inconsolata}

\usepackage{graphicx}

\usepackage{todonotes}
\usepackage{adjustbox}
\usepackage{color}
\usepackage{colortbl}
\usepackage{multirow}
\usepackage{graphicx,wrapfig}
\usepackage{enumitem}
\usepackage{booktabs}
\usepackage{tabularx}
\usepackage{makecell}
\title{Enhancing Rhetorical Figure Annotation: An Ontology-Based Web Application with RAG Integration}

\author{
 \textbf{Ramona Kühn\textsuperscript{1}},
 \textbf{Jelena Mitrović\textsuperscript{1,2}},
 \textbf{Michael Granitzer\textsuperscript{1}}
\\
\\
 \textsuperscript{1}University of Passau
 \textsuperscript{2}Institute for AI Research and Development of Serbia
\\
 \small{
   \textbf{Correspondence:} \href{mailto:ramona.kuehn@uni-passau.de}{\{ramona.kuehn, jelena.mitrovic, michael.granitzer\}@uni-passau.de}
 }
}

\begin{document}
\maketitle

\begin{abstract}
Rhetorical figures play an important role in our communication. They are used to convey subtle, implicit meaning, or to emphasize statements. We notice them in hate speech, fake news, and propaganda. By improving the systems for computational detection of rhetorical figures, we can also improve tasks such as hate speech and fake news detection, sentiment analysis, opinion mining, or argument mining. Unfortunately, there is a lack of annotated data, as well as qualified annotators that would help us build large corpora to train machine learning models for the detection of rhetorical figures. The situation is particularly difficult in languages other than English, and for rhetorical figures other than metaphor, sarcasm, and irony. 
To overcome this issue, we develop a web application called ``Find your Figure'' that facilitates the identification and annotation of German rhetorical figures. The application is based on the German Rhetorical ontology GRhOOT which we have specially adapted for this purpose. In addition, we improve the user experience with Retrieval Augmented Generation (RAG). In this paper, we present the restructuring of the ontology, the development of the web application, and the built-in RAG pipeline. We also identify the optimal RAG settings for our application. Our approach is one of the first to practically use rhetorical ontologies in combination with RAG and shows promising results.
\end{abstract}
\section{Introduction}
\label{sec:intro}
The consideration of rhetorical figures from a computational perspective is important, as they can convey subtle, implicit meanings (e.g., metaphors), make texts more memorable, or add emphasis to the message (e.g., through the repetition of words). 
%because every rhetorical figure has a specific function on the audience~\cite{crocker1977social,harris2018annotation,harris2023rules}. 
Their detection in text regularly leads to improved performance of various NLP applications, such as hate speech~\cite{lemmens2021improving} or fake news detection~\citep{dwivedi2021survey,fang2019self,rubin2016fake,troiano2018computational}, sentiment analysis~\citep{ranganath2018understanding}, or persuasive communication in general~\citep{anzilotti1982rhetorical,gass2022persuasion,ranganath2018understanding}.

Unfortunately, most computational approaches for the detection of rhetorical figures struggle with lower performance than they could actually achieve, e.g., ~\citet{bhattasali2015automatic,dubremetz2015rhetorical,zhu2022configure}.~\citet{kuhn2024elephant} identify the major challenges for researchers in this domain and point out why their approaches often suffer from lower performance. One of the main reasons is the lack of data or unbalanced datasets. In these datasets, the number of instances without rhetorical figures is higher than those containing them, as shown in~\citet{adewumi2021potential,bhattasali2015automatic,dubremetz2017machine,kuhn2023hidden,kuhn2024using,ranganath2018understanding}.

Another major problem is that most detection approaches focus on English. This means that (annotated) data in other languages are even scarcer. In addition, annotators qualified in this research field are not easy to find, and the quality of their annotations varies greatly due to the ambiguous nature of rhetorical figures~\citep{strommer2011using,gavidia2022cats}.

\begin{figure}[!h]
    \centering
    \includegraphics[width=1\linewidth]{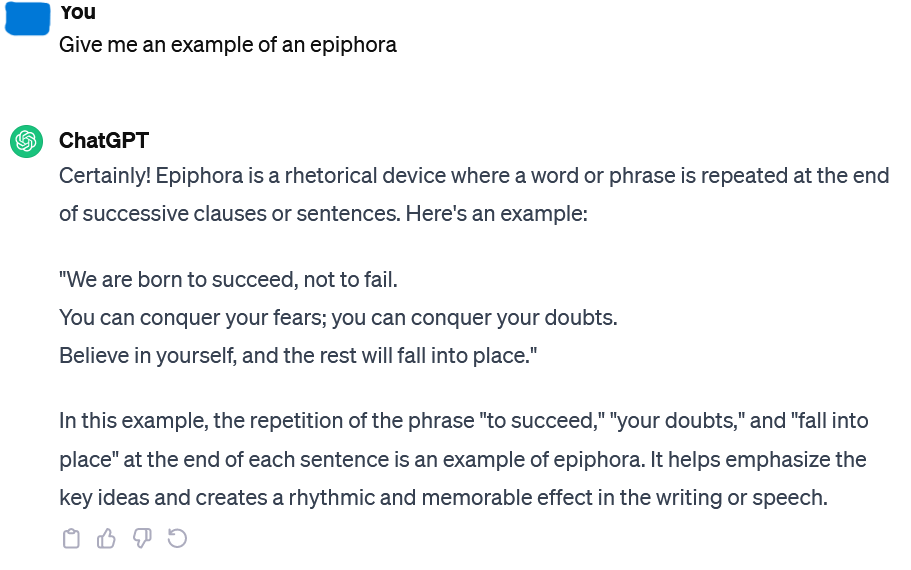}
    \caption{Powerful LLMs such as ChatGPT still fail when asked for examples of certain rhetorical figures. In this example, the LLM claims that the last words in the sentences are repeated.}
    \label{fig:epiphora}
\end{figure}

We overcome those two major challenges in the domain of rhetorical figure detection by developing an interactive web application for the collection of rhetorical figures in German. Users without linguistic knowledge are guided through an interactive process in which they can determine the name of a rhetorical figure in a text and annotate it. In addition to the guided input, we offer a chat-like feature, so users can interact with a large language model (LLM). The LLM obtains domain-specific knowledge about rhetorical figures through retrieval augmented generation (RAG)~\citep{lewis2020retrieval}.
Although current LLMs are powerful, they often struggle even in English with generating and detecting rhetorical figures, especially syntactic figures. An example is shown in Fig.~\ref{fig:epiphora}, where ChatGPT is asked to generate sentences that contain an epiphora. Epiphora is a rhetorical figure, where the last word of each sentence is repeated at the end of the next sentence (e.g., ``When I was a child, I spoke as a child, I understood as a child, I thought as a child''). Although the LLM in our example is aware of the definition of epiphora, it fails to apply it in a sentence. The situation is even worse in languages other than English.

RAG reduces hallucinations and enables LLMs to obtain domain-specific knowledge, making it especially useful in areas where fine-tuning is constrained by limited data. Rhetorical figures are such a field with scarce annotated data~\citep{dubremetz2017machine}. RAG requires only an external knowledge source, often a document. This makes it easier to adapt RAG if the information in the document changes. We use an adapted version of the German GRhOOT ontology of rhetorical figures~\cite{kuhn2022grhoot} as an external source of knowledge. We experiment with different configurations and chunking methods to find the optimal setting for our purpose. We evaluate RAG's performance with the Ragas framework~\citep{es2023ragas} and a ground truth file based on ontological competency questions~\citep{gruninger1995methodology,allemang2011semantic,noy2001ontology,hristozova2002extreme}.
Competency questions are specifications of ontologies ensuring that they have enough knowledge to answer the questions of users. They are usually formulated during the ontology design process.\newline \newline

Our contributions in this paper are as follows:
\begin{itemize}
    \item \textbf{Web Application Development:} We develop a web application called ``Find your Figure'' to overcome the lack of annotated data for rhetorical figures in German.
    \item \textbf{Ontology Restructuring:} We restructure and simplify the German GRhOOT ontology. The simplification of relations and properties is called reification in the context of ontologies. The reified GRhOOT ontology serves as the basis for our web application.
    \item \textbf{LLM Integration with RAG:} We integrate an LLM with RAG to ensure natural interaction with the ontology, and test different settings and chunking methods to find the most effective configuration for our needs.
    \item \textbf{Performance Evaluation:} We evaluate the performance of the integrated RAG LLM using the competency questions of the GRhOOT ontology. 
    %\item We are comparing the usefulness of RAG vs. LLM-generated SPARQL queries (``text-to-SPARQL'').
\end{itemize}

The code and supplementary material are available online.\footnote{The code is available on GitHub: \url{https://github.com/kuehnram/FindYourFigure}.}

% Our overall goal is to facilitate the collection of annotated text containing rhetorical figures. We enable users to search the ontology in natural language with the help of a web application and the assistance of powerful LLMs. If we collect enough data of rhetorical figures through the web application, we can improve the perf
\section{Related Work}
The scarcity of annotated data and the high imbalance between classes with or without rhetorical figures is a well-known issue in the domain of computational detection of rhetorical figures~\citep{dubremetz2015rhetorical,bhattasali2015automatic,dubremetz2017machine,ranganath2018understanding,adewumi2021potential,kuhn2023hidden,kuhn2024elephant}. Unfortunately, efforts to overcome the problem are limited.~\citet{chakrabarty2022flute} use OpenAI's gpt-3 to generate text containing a rhetorical figure, but their method still requires three human annotators to oversee the output. Additionally, annotations of rhetorical figures often have a high variability, because annotators cannot agree on the existence of a figure in particular cases~\citep{strommer2011using, dubremetz2015rhetorical,troiano2018computational}. This problem can be directly linked to the lack of consensus and multiple varying definitions, inconsistent names, and spellings of rhetorical figures, which is a well-known problem~\citep{harris2018annotation,gavidia2022cats,kuhn_2024_10698380}.

Rhetorical ontologies aim to address this issue by building formal models to standardize definitions and descriptions. Important ontologies in this domain are the English RhetFig~\citep{kelly2010toward}, Ploke~\citep{wang2021ontology}, and ESTHER ontology~\citep{kuhn2023esther}, the Serbian RetFig~\citep{mladenovic}, and the German GRhOOT ontology~\citep{kuhn2022grhoot}. However, none of those ontologies has yet been applied in a practical scenario for collecting or annotating rhetorical figures.
%todocite \citet{kuhnmultilingual} suggest different application scenarios.

Our work addresses this gap by implementing a web application based on the German GRhOOT ontology. In addition, we use the ontology to enhance the context of an LLM through RAG. \citet{lewis2020retrieval} show the effectiveness of RAG in different NLP tasks such as question answering and generation, while outperforming pre-trained models.~\citet{zhao2024retrieval} present more domains in which RAG is useful, such as the video, audio, or text domain. 

%As we illustrated, rhetorical figures represent a domain with limited data, especially in languages other than English. RAG is a suitable approach to combine natural interaction with an LLM while reducing its tendency for hallucinations.
\section{Reification of the GRhOOT ontology}
\label{sec:reified}
The German GRhOOT ontology was developed by \cite{kuhn2022grhoot}. It contains the formal description of 110 common German rhetorical figures. Each figure is specified based on the way it is constructed. The figure epiphora shall serve as an example here. In an epiphora, the last word of each sentence is repeated at the end of the next sentence. In the ontology, these properties are expressed by the relations\\
\texttt{:Epiphora :isInPosition :Beginning .} \\
\texttt{:Epiphora :isInArea :Sentence .}\\
\texttt{:Epiphora :isRepeatableElementOfSameForm \\
\centerline{:Word .}}\\
% \texttt{Epiphora $\rightarrow$ isInPosition $\rightarrow$ Beginning} \\
% \texttt{Epiphora $\rightarrow$ isInArea $\rightarrow$ Sentence}\\
% \texttt{Epiphora $\rightarrow$ isRepeatableElementOfSameForm \\
% \centerline{$\rightarrow$ Word}}\\
Rhetorical figures in the ontology also contain relations to express a textual definition, example sentences, and names of the figure in other languages. An example of the complete formal model of epiphora is shown in Fig.~\ref{fig:epiphoraAppendix} in the Appendix in Section \ref{sec:appendix}. While building the web application and specifying user needs, we identify opportunities for further enhancement, particularly in simplifying the relations within the ontology. 
For this reason, we create an adapted version of the GRhOOT ontology.
The main changes are the \textbf{reification} of relations and a more \textbf{fine-grained description} of definitions, authors, example sentences, and their sources. In addition, we model rhetorical figures as \textbf{classes} instead of individuals. Reification involves breaking down properties and relationships into more fine-grained components while offering a more detailed and flexible representation of the ontology. Although this approach increases complexity~\cite{stevens2010reification}, it allows for more precise querying and filtering of attributes. Consider for example the construction relationship \newline 
\texttt{:isRepeatableElementOfSameForm :Word .}

When a user wants to list all figures that contain a repetition of a word, it would be a cumbersome operation to filter the relation names for the \texttt{isRepeatable} substring. We want to make the search more intuitive by breaking compound relations into smaller, fine-grained ones. For example, we split the repetition relation into three relations (RF denotes Rhetorical Figure): \newline
\texttt{:RF :hasOperation :Repetition .} \\ \texttt{:RF :affectedElement :Word .} \\ \texttt{:RF :operationalForm :SameForm .} \\
A comparison between old and new relations is shown in Table~\ref{tab:reificationExample} in Example (a). There were several relations of this form that we adapted accordingly.

This change allows users of the web application to generally filter for figures with the same operations, the same affected elements, or the same operational forms. Reification often makes relations more implicit, requiring an understanding of which relations belong together. However, users are guided through our web interface, so we do not see any drawbacks.

Moreover, we adapt the ontology to reflect the hierarchical structure of rhetorical figures. In the original GRhOOT, rhetorical figures are modeled as individuals, similar to the Serbian RetFig ontology~\cite{mladenovic}. However, the English RhetFig~\cite{kelly2010toward} and the English Ploke ontology~\cite{wang2021ontology} model figures as classes. We also decided to convert the rhetorical figures from individuals to classes to better align with established hierarchies in rhetorical theory~\cite{harris2017rhetorical,o2018arguments}. This adjustment is particularly useful in annotation tasks, where it is important to recognize that one figure may be a more specific form of another figure, and both annotations can be valid. For example, the figure antimetabole is a more specific form of the figure chiasmus and can therefore be modeled as a subclass of chiasmus.

Another major change is the adaption of textual definitions and example sentences that are no longer modeled as property relations.
% For example, a textual definition in the current GRhOOT looks like this 
% \texttt{Anaphora $\rightarrow$ hasDefinition $\rightarrow$ ``An anaphora is the repetition of the first word in consecutive sentences - John Doe.''} \\
% Textual definitions and authors are currently mixed. The same is true for the example sentences of rhetorical figures.
% \texttt{Anaphora $\rightarrow$ hasExample $\rightarrow$ ``Das Wasser rauscht, das Wasser schwoll'' Der Zauberlehrling (Goethe).} \\
%This structure does not provide any standardized form on how examples should be added to the ontology. 
We convert the actual definitions and examples into individuals, called e.g., \texttt{DefinitionAnaphora1}, or \texttt{Example1}. This way, we can add multiple definitions to a rhetorical figure, reflecting the great variety of definitions from different authors and perspectives. Furthermore, by not directly naming examples according to the figure, e.g., \texttt{ExampleAnaphora1} but only \texttt{Example1}, we can reuse the example for other figures, because multiple figures are often co-located. This means that an example can be assigned to the figure anaphora as well as to another related figure (symploke, epiphora, parallelism, etc.). In addition, the new structure reduces redundancy in the ontology. The new construction of the examples is shown in Table~\ref{tab:reificationExample} in Example (b) for the definitions and Example (c) for the textual examples.

\renewcommand{\arraystretch}{1.2}
\begin{table*}[!ht]
\adjustbox{max width=\linewidth}{%
\begin{tabular}{>{\ttfamily}l>{\ttfamily}l>{\ttfamily}l} \hline

\cellcolor[HTML]{EFEFEF} \normalfont{Example} & \cellcolor[HTML]{EFEFEF} \normalfont{Original GRhOOT}  & \cellcolor[HTML]{EFEFEF}\normalfont{Reified GRhOOT}  \\

\normalfont{(a)} & :RF~~:isRepeatableElementOfSameForm~~:Word                                          & 

\begin{tabular}{p{2.5cm}p{3cm}p{7.5cm}}
    
    :RF & :hasOperation & :Repetition ; \\
     & :affectedElement &:Word ;\\
     & :hasOperationForm & :SameForm . \\
    
% \texttt{hasOperation $\rightarrow$ Repetition} \\ \texttt{affectedElement $\rightarrow$ Word} \\ \texttt{hasOperationForm $\rightarrow$ SameForm} 
\end{tabular}

                    \\ \hline
\normalfont{(b)} & :RF~~rdfs:comment~~:Repetition of the first word {[}...{]}                         & 

\begin{tabular}{p{2.5cm}p{3cm}p{7.5cm}}
    :RF & :hasDefinition & :DefinitionRF1 . \\
    :DefinitionRF1 & :hasAuthor & ``Gerd Berner'' ; \\
     & :isDefinition & ``Repetition of the first word {[}...{]}'' .
\end{tabular}
                                                                      \\ \hline
\normalfont{(c)} & \makecell[l]{:RF~~:isExample~~:The water {[}...{]}. The water {[}...{]}\\ (J. W. Goethe, Der Zauberlehrling)} & 

\begin{tabular}{p{2.5cm}p{3cm}p{7.5cm}}
    :RF & :hasExample & :Example1 .     \\
    :Example1  & :hasAuthor & ``Johann Wolfgang von Goethe'' ; \\
    & :hasSource & ``Der Zauberlehrling'' ; \\
    & :isExample & ``The water {[}...{]}. The water {[}....{]} '' .
\end{tabular}

\\ \hline

% \multicolumn{2}{c}{\cellcolor[HTML]{EFEFEF}Rhetorical Figures as Classes instead of Individuals}                                                                                                                                                                                                                                                                                                                                                       &  &  &  \\
% Individual: Anaphora                                                                                       & \begin{tabular}[c]{@{}l@{}}Class: Anaphora\\ Parents: Rhetorical Figure\end{tabular}                                                                                                                                                                                                                                                      &  &  &  \\ \hline
\end{tabular}
}
\caption{Example of reified relations that we changed in the new version of the GRhOOT ontology (RF = Rhetorical Figure).}
\label{tab:reificationExample}
\end{table*}

Fig.~\ref{fig:epiphora_new_grhoot} in the Appendix in Section~\ref{sec:appendix} shows an example of the figure epiphora in the reified GRhOOT ontology.

\section{Ontology-Based Web Application for Rhetorical Figure Annotation}
\label{sec:webapp}
Our overall goal is to improve the computational detection of rhetorical figures by collecting more annotated instances of rhetorical figures.
%make the rhetorical ontology GRhOOT accessible for a wider range of users, and to collect annotated instances of rhetorical figures in German. 
We demonstrated the important role of rhetorical figures and how their detection can improve many NLP systems in Section~\ref{sec:intro}. The interaction with the ontology through the web application is as natural and intuitive as possible without the need for linguistic knowledge or knowledge about ontological details. When users encounter a sentence in which they suspect a rhetorical figure, they can use our web application to determine its name and function. The application is based on the Python Flask\footnote{\url{https://flask.palletsprojects.com/en/3.0.x/}} framework and uses an SQLite~\footnote{\url{https://www.sqlite.org/}} database. The Flask framework is suited for lightweight web applications such as ours. 

\subsection{Pages of the Web Application}
\begin{figure}
    \centering
    \includegraphics[width=\linewidth]{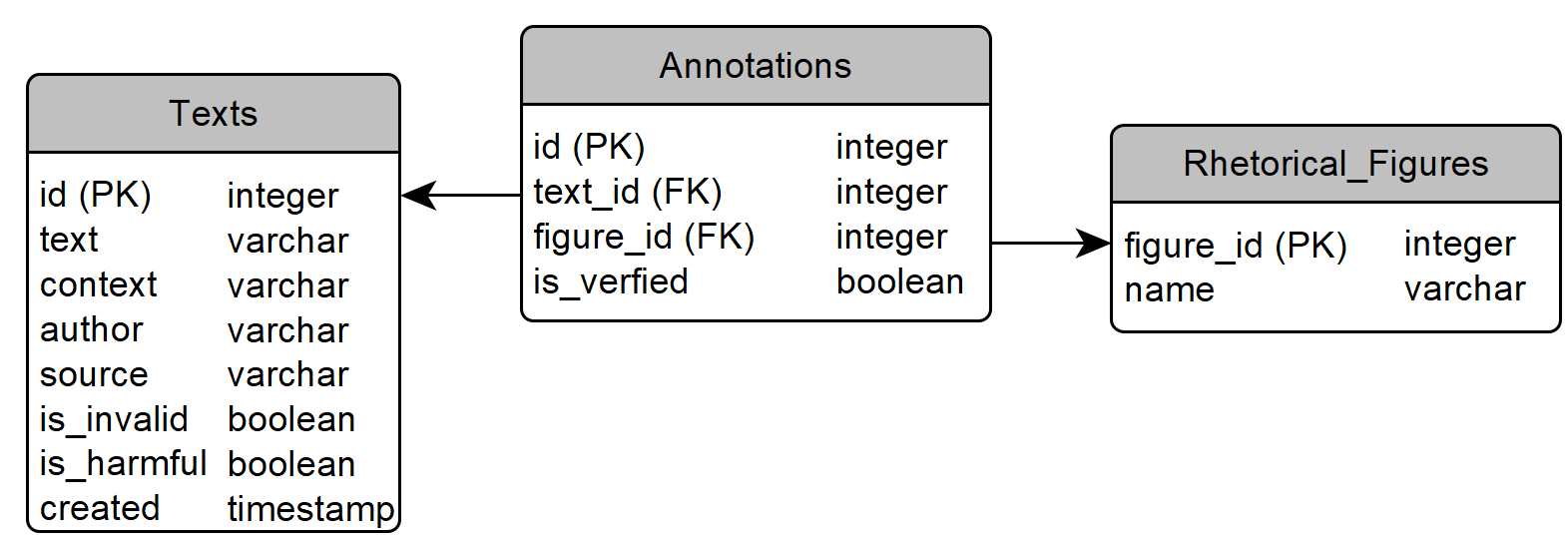}
    \caption{Scheme of the SQL Lite Database. The arrows indicate foreign key (FK) relations. PK denotes primary keys.}
    \label{fig:dbscheme}
\end{figure}

\begin{figure*}[!h]
\centering
    \includegraphics[width=0.6\linewidth]{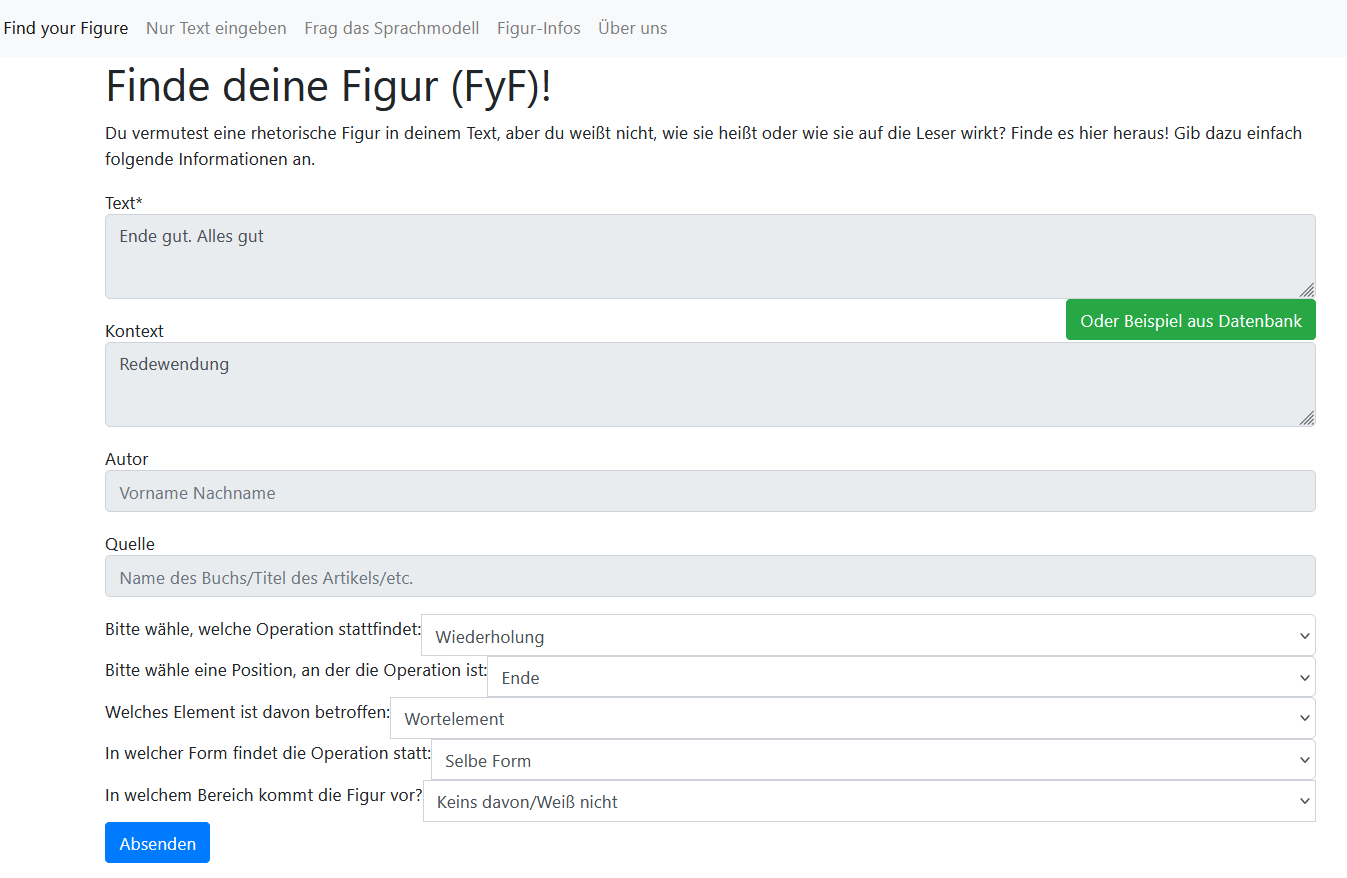}
    \caption{The page \texttt{FyF.html} helps users to find the name of a rhetorical figure hidden in a text. Properties can be selected from the dropdown lists.}
    % \Description[Overview of the FyF.html page]{The page shows text fields and drop down fields where users can select the properties of the text they found.}
    \label{fig:fyf}
\end{figure*}

\begin{figure}[!h]

    \includegraphics[width=\columnwidth]{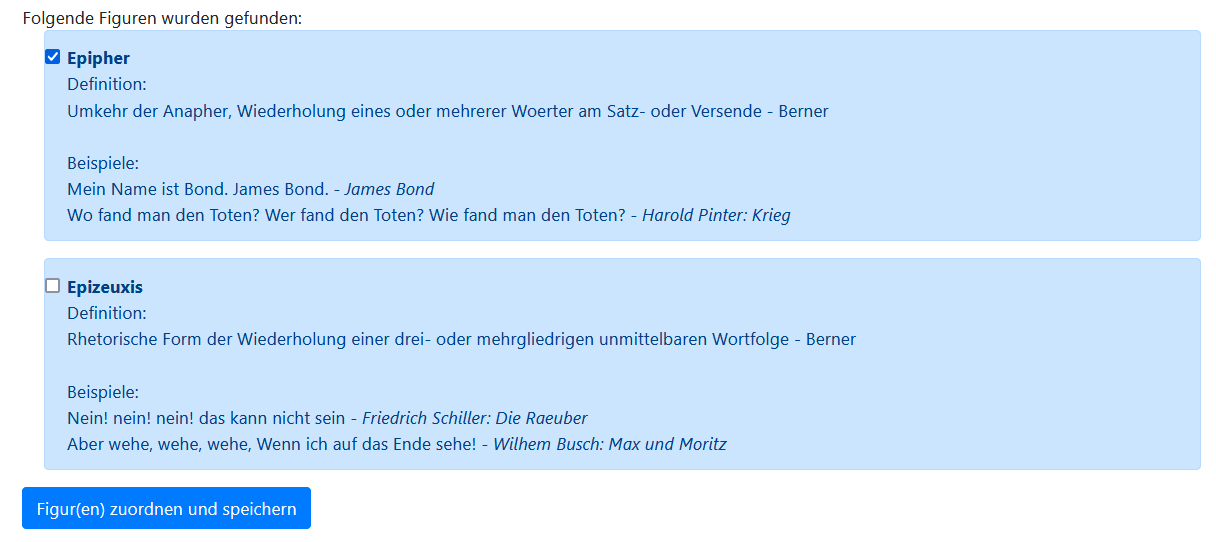}
    \caption{The result of the user's submission. Rhetorical figures fulfilling the properties are displayed along with the option to select suitable ones.}
    % \Description[Overview of the result of a user query]{Two matching rhetorical figures are presented to the users. The users can select in a box which example they want to associate with the text.}
    \label{fig:annotate}
\end{figure}

The application encompasses the following five pages.
\begin{itemize}
    \item \texttt{create.html}: On this page, users have the possibility to enter a sentence with a rhetorical figure without annotating it. Users submit a text or sentences, context (e.g., preceding sentences, description of the situation), author, and source of the text. The example is stored in the database for later annotation by other users who do not have an own example but choose a random one from the database. %The page is shown in Fig.~\ref{fig:create}.
    \item \texttt{FyF.html}: This is the main page of our ``Find your Figure'' application. It is shown in Fig.~\ref{fig:fyf}. Users choose to submit their text/sentence or choose a random one from the database previously submitted by users on the \texttt{create.html} page.
    The option to enter an own text also includes specifying context, author, and source.
    The users then select the properties of the text from a dropdown list that best describes the pattern in the submitted text. Properties are extracted relations from the ontology, such as \texttt{operation} (e.g., \texttt{repetition}), \texttt{affected element} (e.g., \texttt{word}). Users always have the possibility to choose \texttt{No idea} (\texttt{Keines davon/Weiß nicht}) if they are not sure about the property. After the users submit the information, the properties are translated into a SPARQL query in the backend and executed on the ontology. If matching figures are found, they are presented along with a definition and examples of the figure in the frontend. The users can choose one or more figures they consider appropriate as shown in Fig.~\ref{fig:annotate}. The text, context, author, source, and annotated figure are then written to the SQLite database. The database scheme is shown in Fig.~\ref{fig:dbscheme}.   
    \item \texttt{llm.html}: As the annotation process in \texttt{FyF.html} still requires basic knowledge of linguistic concepts, we integrate a chatbot-like feature for a more natural interaction between users and ontology. Users simply submit the example text they want to annotate and describe its properties to the LLM. %This page is shown in Fig.~\ref{fig:llm}.
    It offers a field for text, context, author, and source, or the possibility to load an example from the database. Instead of the drop-down list, a text field is presented for the LLM prompt. The answers are generated by the LLM with RAG extended context. The different setups to find the best RAG parameters are described in Section~\ref{sec:ragOntology}. 
    \item \texttt{figure\_info.html}: This page provides an informative overview of rhetorical figures. Users can select the name of a rhetorical figure from a dropdown list. The application then presents definitions of the figure and example sentences. As the elements of the list and their information are retrieved from the ontology, it can be easily extended by adapting the ontology.
    \item \texttt{about.html} This page presents our research project and offers an imprint with contact details.
\end{itemize}

% \begin{figure}
%     \centering
%     \includegraphics[width=\linewidth]{figs/create.PNG}
%     \caption{The page \texttt{create.html} lets users submit an example without annotating it.}
%     \Description[Overview of create.html]{The page shows texfields and a submit button.}
%     \label{fig:create}
% \end{figure}

% \begin{figure}[!h]
%     \includegraphics[width=\columnwidth]{figs/llm.PNG}
%     \caption{The page \texttt{llm.html}, where the users have the possibility to ask the RAG extended language model about rhetorical figures.}
%     % \Description[The page llm.html]{Texfields for sentences, authors, sources, and a text field to pose the questions to the LLM.}
%     \label{fig:llm}
% \end{figure}

% The specificities of the figures that users can choose from drop-down menus are directly extracted from the ontology. From this information about the figure, a SPARQL query is built in the backend and executed on the ontology. If figures are found that fulfill all the properties, they are presented to the users, along with a definition of the figure and examples. The users can choose one or more figures they consider matching. Those examples are then written to the SQLite database. 
% In this database, we have a table for the submitted texts, a table for the rhetorical figures, and a table connecting the texts with the figures via foreign keys.

% As this annotation process still requires a basic knowledge of linguistic concepts, we integrate a chat a chatbot-like feature for a more natural interaction between users and ontology. Users simply submit the example text they want to annotate and describe its properties to the LLM.

Annotated text submitted by the users can be added to the ontology as examples of the respective figure. However, we did not implement this functionality yet, as we first want to verify that the submitted examples are correct.

\subsection{Verification of User Input}
We need to verify that
\begin{enumerate}
    \item users do not violate intellectual property rights when uploading examples,
    \item the submitted text is valid and not ``gibberish'',
    \item the assigned rhetorical figures is correct,
    \item the submitted text is not harmful or violating, especially when presented to other people for annotation.
\end{enumerate}

It is important to verify that no intellectual property rights are infringed, especially when we later train models on the obtained data. 
Researchers are aware of those challenges regarding intellectual property rights when training LLMs with large amounts of text~\citep{smits2022generative}. However, identifying unauthorized material or a violation of intellectual property is ``notoriously difficult to prove''~\citep{chesterman2024good}. To overcome the first challenge (1.), users must indicate at least an author or source of their submitted text. In addition, we raise awareness of the problem by displaying an informational text next to the author and source fields in the \texttt{create.html} and \texttt{FyF.html} pages. The LLM we are using in the \texttt{llm.html} page is used in a generative way, but we prompt it with information from the ontology in a RAG setting to ensure that it is more likely to only present examples from the ontology. Another possibility to make potential copyright infringements tractable is user authentication via login. However, this would increase the threshold to use the app, especially for younger users, e.g., school children. 
% However, we use LLMs not in a generative but more in a discriminative way

Furthermore, we want to verify that users only submit valid text which is then written to the database (2.). We cannot rely (solely) on common grammar or spell checks, as odd grammatical constructs or omitted letters can be a feature of rhetorical figures. 
We use a combination of a language detector (\texttt{langdetect}) to identify if the submitted example is German, a measure of the text length (10 $\leq$ \texttt{text\_length} $\leq$ 1000, and a grammar checker that supports German (\texttt{language-tool-python}\footnote{\url{https://pypi.org/project/language-tool-python/}}). If one of those three checks fails, we ask gpt-3.5-turbo to evaluate if the text is ``gibberish''. If the answer is positive, we show a notification to the users if they really want to submit the example, making them aware of potential problems. If they submit the example anyhow, we will flag the example in the database in the column \texttt{is\_invalid}, such that an administrator can later check the validity. 

Ensuring that the rhetorical figures assigned by the users are correct (3.) is a highly challenging task. A different detection algorithm would be required for each figure. However, especially for figures other than metaphor, irony, and sarcasm, if any approaches exist, they often have lower performance~\citep{kuhn2024computational}. In addition, each approach is language dependent and often requires high manual efforts to achieve acceptable performance~\cite{mladenovic2017using}. Using existing lexical resources for rule-based approaches, such as German wordnets, is barely an option as they do not contain enough information for a reliable detection~\citep{kuhn2023hidden} and are in general difficult to maintain~\citep{mladenovic2014developing}. Even approaches based on language models are difficult to implement as data is too scarce or often too imbalanced for training models~\citep{dubremetz2015rhetorical,kuhn2024using}. However, we use a rule-based check if a figure of perfect lexical repetition is assigned and verify that at least two words are repeated in the same form. Nevertheless, we still have to rely on manual checks by an administrator for all figures. For this purpose, the column \texttt{is\_verified} in table \texttt{Annotation} (see Fig.~\ref{fig:dbscheme}) is intended to mark if an example has already been approved.

Users can annotate random examples from the database on the \texttt{FyF.html} page. However, we want to ensure that potentially harmful content is not presented (4.), especially when the web application is used by young children who learn about rhetorical figures. We do not prevent per se the submission of harmful content, as certain rhetorical figures occur frequently in hate speech, e.g., sarcasm~\citep{frenda2023sarcasm}. However, we exclude those examples to be shown to other users. We introduce the boolean field \texttt{is\_harmful} to mark those examples and prevent their retrieval for annotation in the \texttt{FyF.html} page. We have not yet implemented any hate speech detection mechanism yet. However, we plan to run a daily check on the database.

% For example, the book title ``Der satanarchäolügenialkohöllische Wunschpunsch'' 

% title of one of Bach's cantata ``Weichet nur, betrübte Schatten!'' would be classified as wrong grammar by the TextBlob package and misspelled by the \texttt{pyspellchecker}.

% \subsection{Admin Control Panel}

% to ensure that the annotation is correct and also confirmed by others. 
% We consider it future work to implement this feature. %It is possible in the future to implement that an instance with enough similar annotations is added to the ontology.

\section{RAG Integration: Parameter Testing and Evaluation}
\label{sec:ragOntology}

\begin{figure*}[!h]
    \centering
    \includegraphics[width=\textwidth]{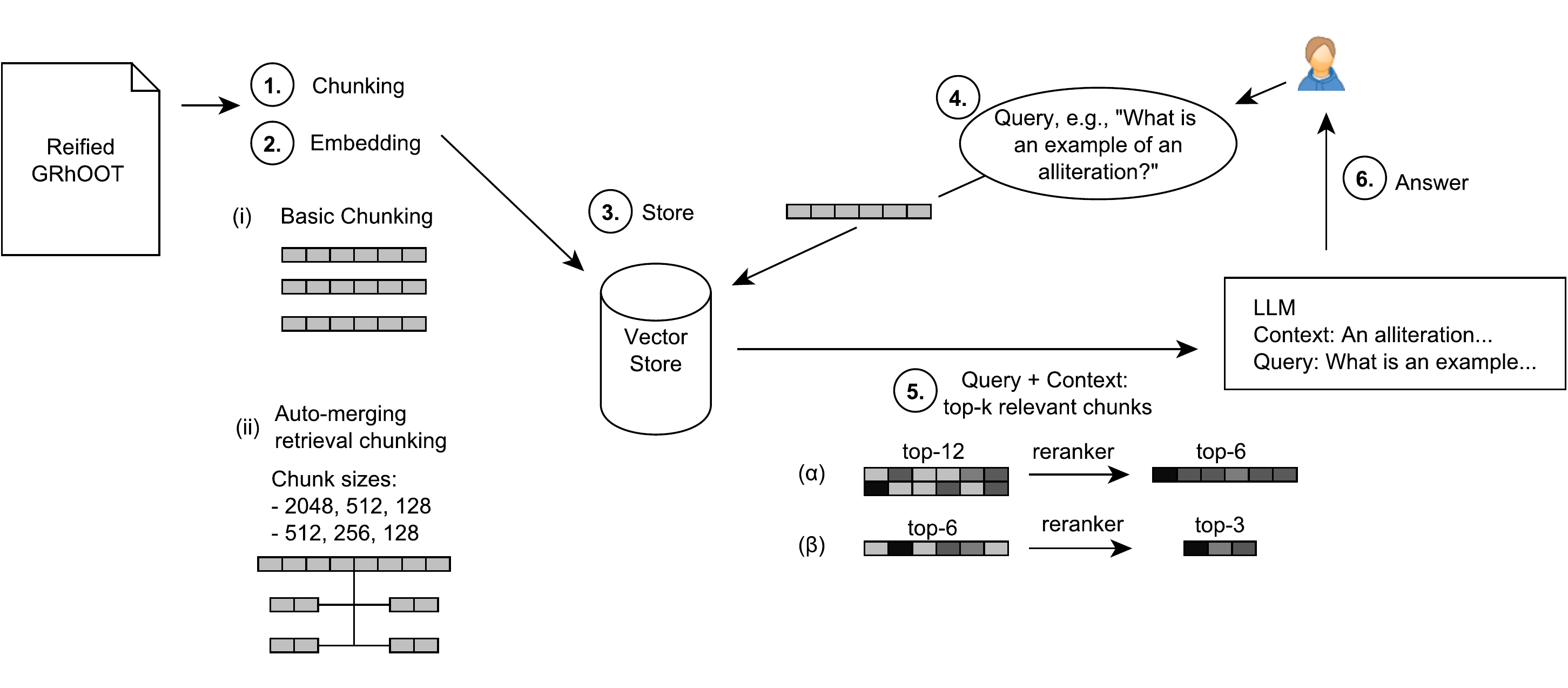}
    \caption{Overview of the integrated RAG Pipeline for the GRhOOT ontology.}
    \label{fig:rag}
     % \Description[RAG Pipeline]{The RAG pipeline shows how RAG works. In this case we also present the different setups that we apply, such as different chunking techniques, different chunk sizes, and varying top-k elements.}
\end{figure*}
% We will compare and LLM with RAG extended context with an LLM-based text-to-SPARQL approach.
RAG uses an external knowledge source to enhance the context of an LLM. 
% Although current LLMs are powerful, they often struggle with generating and detecting rhetorical figures, especially syntactic figures in languages other than English. 
LLMs still struggle to generate rhetorical figures as we showed in Fig.~\ref{fig:epiphora}. Another challenge in this domain are the many different and varying definitions of rhetorical figures. Though, we want the integrated LLM to respond to the web application users with the specific definitions we use in our ontology to ensure consistent annotation guidelines.

Fig.~\ref{fig:rag} illustrates the RAG pipeline with its individual steps. It also shows the different parameters used in our experiment to determine the optimal settings for the web application. The input is the reified GRhOOT ontology which is also the basis of the web application. After chunking and embedding the information (steps 1 and 2), it is stored in a vector store (step 3). When a user asks the system a question about rhetorical figures in step 4 (e.g., ``What is an alliteration?''), the question is embedded and compared to the content in the vector store. The top-k relevant chunks are retrieved in step 5, often in combination with a reranker, and added as context to the query, e.g., actual examples of an alliteration. The LLM receives the context along with the original question. In step 6, the LLM answers the users' questions with a reduced probability of hallucinations and knowledge of the domain, which are rhetorical figures in our case.

% We create an application ontology out of the GRhOOT ontology for better adaptability. This application ontology consists only of 15 instead of 110 rhetorical figures, making changes more feasible for different scenarios. For our RAG approach on the ontology, we test different document structures (a-c).
% \begin{enumerate}[label=(\alph*)]
%     \item \textbf{Basic GRhOOT}: We use a shorter version of the original GRhOOT ontology to perform RAG.
%     \item \textbf{Reified GRhOOT}: We use a restructured, reified version of the shorter GRhOOT ontology. This means that relations and connected elements are dispersed across the file (e.g., separating the definition of a rhetorical figure from its construction rules and property descriptions).
%     \item \textbf{Continuous GRhOOT}: We reformulate the shorter GRhOOT ontology into continuous text with the help of gpt-3.5-turbo. As RAG is mostly applied to and optimized for continuous text, we assume that RAG performs better on continuous text than on semi-structured formats such as ontologies.
%     % \item We use a basic chunking technique and compare it to an auto-merging retrieval strategy. In auto-merging retrieval, chunks are divided into a tree-like structure with smaller leaf chunks. If relevant information is found in a leaf chunks, the parent chunk is returned.
% \end{enumerate}

To find the best setting, we experiment with different chunk sizes ([2048],[2048, 512, 128], and [512, 256, 128]). It is a known phenomenon called ``lost in the middle'' that content stored in the middle of a chunk is more difficult for the LLM to recall~\cite{liu2024lost}. Therefore, we test different sizes to avoid this problem. As chunking technique, we investigate basic chunking (i) and auto-merging retrieval chunking (ii) with the \texttt{HierarchicalNodeParser} by LlamaIndex.\footnote{\url{https://docs.llamaindex.ai/en/stable/api_reference/node_parsers/hierarchical/}} It is an advanced chunking technique, where a smaller chunk that contains relevant information is merged into the parent chunk and provided as context.

% All settings (a-c) are tested both with basic chunking (i) (chunk size of 2048) and advanced, auto-merging retrieval chunking (ii). For this, we use the \texttt{HierarchicalNodeParser} by LlamaIndex with chunk sizes of [2048, 512, 128] and [512, 256, 128] to see how different sizes affect the results.

We use the multilingual model bge-m3\footnote{\url{https://huggingface.co/BAAI/bge-m3}} for the embeddings.

Usually, the vectorized index is stored in a suitable vector database, which offers additional benefits, especially when working with multiple documents. However, as we only focus on a compact and relatively small ontology, we do not need a vector database. Instead, we store the index in a file. This approach reduces the number of variables that could affect the output, as we are not relying on optimization strategies of database vendors.

\begin{table*}[!ht]

\adjustbox{max width=\linewidth}{%
\begin{tabular}{llclrrrrrr} \toprule
Document & Chunk Sizes                                                    & Chunking Method                       & reranker                        & faithf.                   & c\_precision                   & c\_recall                      & a\_correctn.                 & a\_similarity                  & a\_relevancy                   \\ \midrule

& {\color[HTML]{000000} 2048}                                   &                 Basic          & top-12/6                         & \textbf{0.9023}                         & 0.5548                         & \textbf{0.9857}                         & 0.7355                         & 0.8655                         & \textbf{0.9489}                                                 \\
& \cellcolor[HTML]{EFEFEF}{\color[HTML]{000000} 2048}           & \cellcolor[HTML]{EFEFEF} Basic  & \cellcolor[HTML]{EFEFEF}top-6/3  & \cellcolor[HTML]{EFEFEF}0.8342 & \cellcolor[HTML]{EFEFEF}0.8496 & \cellcolor[HTML]{EFEFEF}0.9098 & \cellcolor[HTML]{EFEFEF}0.5678 & \cellcolor[HTML]{EFEFEF}\textbf{0.9673} & \cellcolor[HTML]{EFEFEF}0.7119  \\
& {\color[HTML]{000000} 2048, 512, 128}                         & AMR                         & top-12/6                         & 0.8760                         & 0.5762                         & 0.9714                         & 0.6481                         & 0.8542                         & 0.8616                                              \\
& \cellcolor[HTML]{EFEFEF}{\color[HTML]{000000} 2048, 512, 128} & \cellcolor[HTML]{EFEFEF}AMR & \cellcolor[HTML]{EFEFEF}top-6/3  & \cellcolor[HTML]{EFEFEF}0.7612 & \cellcolor[HTML]{EFEFEF}0.9009 & \cellcolor[HTML]{EFEFEF}0.9254 & \cellcolor[HTML]{EFEFEF}\textbf{0.8619} & \cellcolor[HTML]{EFEFEF}0.6625 & \cellcolor[HTML]{EFEFEF}0.8041  \\
& {\color[HTML]{000000} 512, 256, 128}                          & AMR                         & top-12/6                         & 0.8817                         & 0.8446                         & 0.6571                         & 0.8311                         & 0.7099                         & 0.8889                                               \\
\multirow{-6}{*}{\begin{tabular}[c]{@{}l@{}}Reified \\ GRhOOT\end{tabular}}                        & \cellcolor[HTML]{EFEFEF}{\color[HTML]{000000} 512, 256, 128}  & \cellcolor[HTML]{EFEFEF}AMR & \cellcolor[HTML]{EFEFEF}top-6/3  & \cellcolor[HTML]{EFEFEF}0.8622 & \cellcolor[HTML]{EFEFEF}\textbf{0.9230} & \cellcolor[HTML]{EFEFEF}0.9190 & \cellcolor[HTML]{EFEFEF}0.5969 & \cellcolor[HTML]{EFEFEF} 0.9608 & \cellcolor[HTML]{EFEFEF}0.5587 \\  \midrule

%%%%%%%%%%%%%%%%%%%%%%%%%%%%%%%%%%%%%

\end{tabular}
}
\caption{Results of our RAG experiments on the reified ontology with different settings. AMR stands for automerging retrieval, an advanced chunking technique. The best scores per column are highlighted in bold.}
\label{tab:evaluation}
\end{table*}

Furthermore, we use a reranker (BAAI/bge-reranker-large) in all settings to improve the results. 
%\footnote{\url{https://www.pinecone.io/learn/series/rag/rerankers/}}. 
In setting ($\alpha$), we first retrieve the top-12 chunks, while the reranker selects the top-6 chunks. In setting ($\beta$), we retrieve the top-6 chunks, while the reranker selects the top-3 chunks ($\beta$).

We are using OpenAI's gpt-3.5-turbo as LLM, as it shows good results in German. We set the temperature to 0.1 to obtain stable responses.

\subsection{Evaluation Setup for the RAG pipeline}
RAG evaluation still poses a challenge. We use the Ragas\footnote{\url{https://github.com/explodinggradients/ragas}} framework for evaluation. It requires a file of questions, answers generated by the LLM in the last step of the RAG pipeline, context information from the original document, and ground truth answers. Most approaches rely on LLM-generated ground truths that are then again evaluated by an LLM.

% In most scenarios, this file is mainly generated by two LLMs, one generator LLM and one critic LLM\footnote{\url{https://docs.ragas.io/en/stable/getstarted/testset_generation.html}}. However, as LLMs suffer from hallucinations, the actual ground truth or context can be false.

Ontologies in general and especially the GRhOOT ontology have a big advantage here. We can use ontological competency questions (CQ)~\cite{gruninger1995methodology,noy2001ontology,hristozova2002extreme,allemang2011semantic} their respective answers extracted from the ontology to generate the ground truth file for the Ragas evaluation. Unfortunately, the GRhOOT ontology comes only with five CQs. However,~\citet{alharbi2023experiment} and~\citet{ciroku2024revont} demonstrate that OpenAI gpt-4 can generate competency questions after the ontology has been created. Therefore, we also use gpt-4 to generate further CQs for the reified GRhOOT ontology. In contrast to~\citet{alharbi2023experiment}, we skip the triple extraction and provide only the formalization of one rhetorical figure as context to the LLM. The LLM is then able to formulate appropriate questions. Additionally, we create template questions asking for properties of rhetorical figures, i.e., ``What is \textit{<property>} of the rhetorical figure \textit{<figure\_name>}'', e.g., ``What is an \textit{example} of the rhetorical figure \textit{anaphora}?''. With those methods, we obtain 70 CQs. We formulate matching SPARQL queries to retrieve the answers from the ontology. The context is manually extracted from the ontology. From this information, we construct our ground truth file in the required format for the Ragas framework. The answers of the LLM still require post-processing before we can use them in the Ragas framework as the LLM tends to add additional quotation marks around figure names or examples but often does not close them. The post-processing step is only necessary for the evaluation. It is not required in the production mode of our web application.

% As this step is still labor-intensive and time-consuming, we create an application ontology using only 15 figures from the reified ontology. This allows us to test different setups more quickly and thoroughly by asking relevant questions for each figure. Although this approach seems initially limited, it is generalizable to the complete ontology. This is an intermediate step to determine the best setup for the entire ontology.

For the evaluation, we use the pre-defined metrics from the Ragas framework. A detailed description can be found online,~\footnote{\url{https://docs.ragas.io/en/latest/concepts/metrics/index.html}} but we describe them shortly here for clarification. We choose the following Ragas metrics:
\begin{itemize}
    \item \texttt{Faithfulness}: Describes the extent to which the answer is grounded in the context. %Verification-function-based calculation.
    \item \texttt{Context precision}: Measures if relevant chunks are ranked higher.% Function-based calculation.
    \item \texttt{Context recall}: Measures if the retrieved context is present in the ground truth.% Function-based calculation.
    \item \texttt{Answer relevancy}: Measures the relevancy of the answer to the question by calculating mean cosine similarity between the actual question and artificial questions generated by an LLM based on the answer.% Function- and LLM-based metric.
    \item \texttt{Answer correctness}: Measures how semantically and factually similar the answer is to the ground truth.% Function-based calculation.
    \item \texttt{Answer (semantic) similarity}: Measures the semantic similarity between the ground truth and the LLM's answer based on the cosine similarity of the embeddings.% Function-based calculation.
\end{itemize}
%answer relevance and context relevance based on LLM evaluation. 
From a user's perspective, the performance of answer metrics, particularly \texttt{answer correctness}, is more important than the performance of context metrics. Therefore, we focus especially on these metrics.

% For our case, it is especially important that the ground truth is literally contained in the answer. Therefore, we introduce a new metric called \textbf{\texttt{answer average match}}. This metric is suited for ontologies as the competency questions often yield only a very short, specified answer without additional descriptive text. For each word in the ground truth, we check that it is contained in the answer of the LLM. We remove both punctuation marks and stopwords from the ground truth. For the stopwords, we use the German stopwords list of the \texttt{nltk.stopwords} package. As we have both longer sentences (e.g., ``the definition of an alliteration is ...'') and single words (e.g.,``Verse'') in our ground truth file, we count the number of words that appear in the LLM's answer. This percentage then indicates the \texttt{answer average match} ranging between 0 and 1.
% However, we faced the problem that German is a language with a high number of inflectional forms. Therefore, the LLM's answer also contains changed inflections such as ``Semantisch'' in the answer instead of ``Semantik'' in the ground truth. As those cases are rare in our files, we added the alternative spelling manually in the check. But this fact has to be kept in mind when applying this metric.

\subsection{Evaluation Results}
In the first review of the answers, we notice that the LLM sometimes responds in English instead of German. For this reason, we add to the prompt the request to only answer in German (step 4 in Fig.~\ref{fig:rag}): ``Bitte antworte nur auf Deutsch!'' (``Please answer only in German!'').

Table~\ref{tab:evaluation} shows the result for the different settings. Surprisingly, advanced chunking techniques do not increase the performance. As best setting, we identify basic chunking with a chunk size of 2048, where the top-12 chunks are selected first and then filtered for the top-6 by the reranker. Only the context precision is low in this setting. However, as we mentioned, we focus more on answer metrics. When reviewing the results, we notice deviations in \texttt{answer correctness} and \texttt{answer similarity}, even when the LLM's answer is correct and semantically similar to the ground truth. We identify the LLM's circumscription of the answer as the problem. For example, consider the following case: \\ \\
\textbf{Question:} ``What is the name of the figure where the first letter of each word sounds the same?'' \\
\textbf{Ground truth:} ``Alliteration''\\
\textbf{LLM's answer }(translated from German): ``The name of the figure where the first letters of each word sounds the same is 'Alliteration'''\\ \\
This answer leads to a reduced \texttt{answer correctness} and \texttt{answer similarity} because of the wordiness of the LLM's answer. However, we do not consider this as an issue since the answer is correct and users would probably prefer a complete sentence over a single-word-response. 

Other examples where the LLM is correct but creates reduced \texttt{answer correctness} and \texttt{answer similarity} are a different choice of words. For example, the LLM answers with this definition for a rhetorical figure \\ \\
\textbf{LLM's answer:} ``Eine Wiederholung des Anfangslautes benachbarter oder nah beieinanderstehender Worte in einem Satz oder Vers'' \\
\textbf{Ground truth:} ``Gleichklingender Anlaut der betonten Silben innerhalb einer Wortgruppe'', \\ \\
where ``Gleichklingender Anlaut'' means the same as ``Wiederholung des Anfangslauts''. We assume that it is caused by the definitions in the ontology taken from dictionaries and books, while the LLM uses more ``modern'' language and simpler expressions.

We see problems when the LLM is asked to answer questions that require aggregation of information (e.g., ``What are the linguistic groups defined in the ontology?''''), which requires reasoning over multiple chunks. This finding is in line with the experience of~\citet{alharbi2023experiment} that LLMs generally tend to fail in such tasks. Unfortunately, RAG can do little to change this. Nevertheless, the high values of the metrics show the efficiency of RAG on ontologies. Even without special adaptions to the ontological structure, we achieve satisfying results.

\section{Conclusion}
The development of the web application to collect rhetorical figures is an important step to overcome the scarcity of annotated data in the field of the computational detection of rhetorical figures, especially in German. The web application allows users to specify the properties of a text and assists them to name and annotate the rhetorical figure hidden in it. Furthermore, the web application serves as an information collection about rhetorical figures, where users can learn the definitions and see examples. The web application, which is built on top of the GRhOOT ontology, is one of the first approaches to practically use a rhetorical figure ontology for figure annotation. It also contains verification functions for the user input. The integrated RAG pipeline allows users to use an LLM-powered chat for the interaction with the GRhOOT ontology. 
One of our main objectives for the future is to publish the web application and promote it to potential users. We can then evaluate the features of the application that are most beneficial to users and learn about their behavior, for example, if they prefer the chat function with the integrated RAG model or the structured drop-down fields.

% We will also collect user feedback and assess the correctness of the annotation of rhetorical figures.
In addition, we will observe the performance of the RAG pipeline. When we collect more examples from users through the web application, it is possible to add them to the ontology and update the vector store to improve the performance of the RAG pipeline. We also envision gamification elements and user sessions to store their achievements to keep them engaged. In addition, we will extend our verification methods for the user input. We also plan to show retrieved chunks to the users so they can compare the information from the ontology with the LLM's answer.

Nevertheless, the current version that combines annotation capabilities and educational resources makes our application a valuable tool in the domain of computational detection of rhetorical figures, as well as a possible interactive resource in education.
%caching

\section{Limitations} 
Our web application for identifying rhetorical figures has some limitations. It is better suited to identify figures with obvious rhetorical features, e.g., figures with repeating elements, than for figures relying on transferred meanings, such as metaphors. However, we see this rather as a limitation on the side of the users. For most persons without linguistic knowledge of rhetorical figures, it is easier to spot and describe obvious lexical patterns than figures with implicit, transferred meaning. Additionally, this app represents only the initial implementation of our envisioned tool. There are many possible features that can be implemented, enhanced, and improved in the future.%&, such as an admin control panel and verification functions for submitted examples to improve accuracy and user experience.

\section{Ethical Considerations}
Regarding the web application, the main concern is the violation of intellectual property rights. Users may submit text from sources they do not have the right to use. Furthermore, the text is then stored in the database and used to train models, even if the original authors did not agree on the distribution of their text. This is not an easy task to solve both from a computational and legal perspective. However, we established methods to encourage users to indicate an author or source of the examples.

Regarding the RAG pipeline, users should be aware that the LLM may produce incorrect answers. We hope to support the users in assessing the truth with our web application, as they can browse the figures to learn about them and with the planned feature to show the retrieved chunks along with the answer of the LLM. This way, users can get a clearer picture if the LLM's answer is correct. 

In conclusion, while RAG on rhetorical ontologies holds significant potential for advancing NLP, addressing these ethical concerns is important to ensure their responsible and fair application.

\section*{Acknowledgements}

% \begin{wrapfigure}{r}{2.5cm}
% \includegraphics[width=2.5cm]{figs/BMBF.jpg}
% \end{wrapfigure}

\begin{figure}[!h]
    \centering
    \includegraphics[width=4cm]{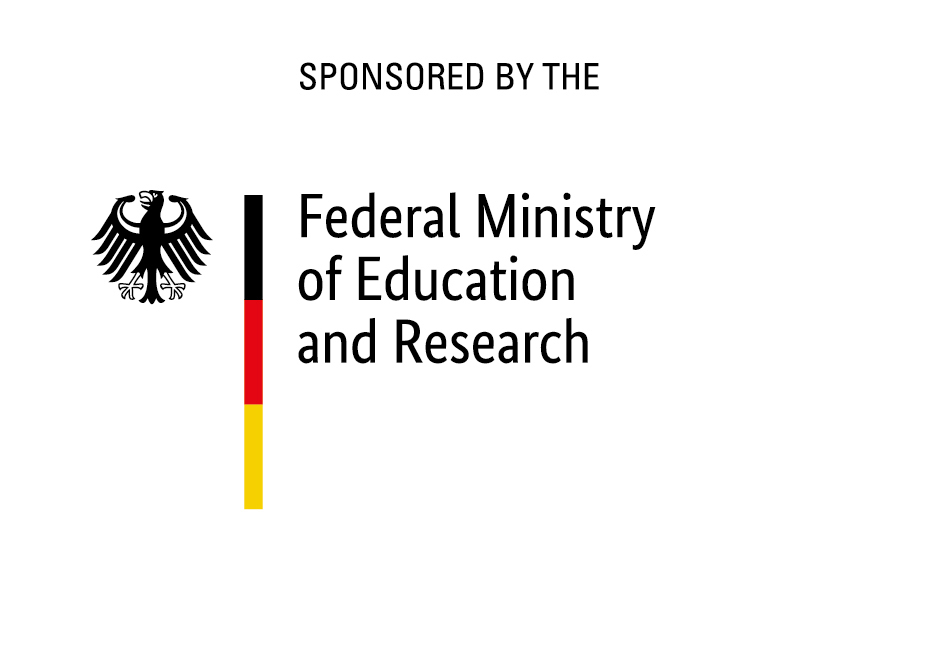}
    % \caption{Caption}
    % \label{fig:my_label}
\end{figure}

The project on which this report is based was funded by the German Federal Ministry of Education and Research (BMBF) under the funding code 01|S20049. The authors are responsible for the content of this publication.

%\input{ack}

\bibliography{custom}

\appendix

\section{Appendix}
\label{sec:appendix}

\begin{figure*}[!h]
    \centering
    \includegraphics[width=1\linewidth]{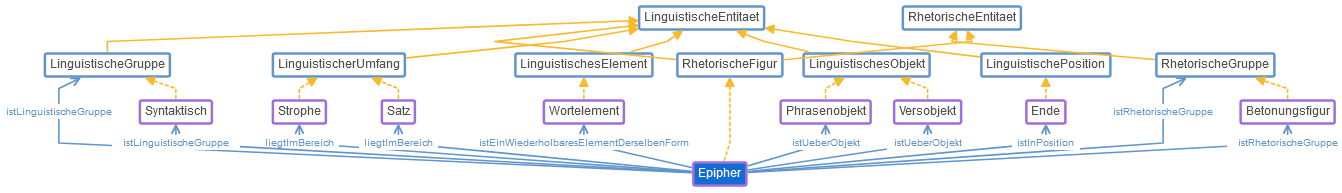}
    \caption{The formal model of an epiphora in the GRhOOT ontology, illustrating the relations between the construction properties.}
    \label{fig:epiphoraAppendix}
\end{figure*}

\begin{figure*}[!h]
    \centering
    \includegraphics[width=1\linewidth]{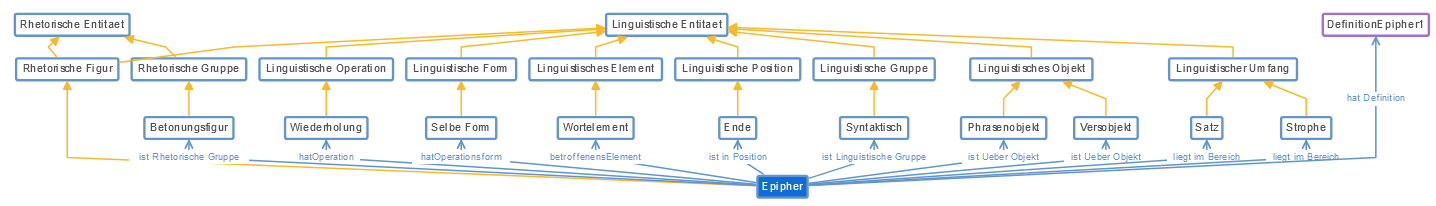}
    \caption{The formal model of an epiphora in the reified GRhOOT ontology. Compared to the model in the original GRhOOT in Fig.\ref{fig:epiphoraAppendix}, relations are simplified. In addition, the concepts are now modeled as classes, illustrated by blue boxes instead of purple boxes that represent individuals.}
    \label{fig:epiphora_new_grhoot}
\end{figure*}

Fig.~\ref{fig:epiphoraAppendix} shows the graphical illustration of the figure epiphora in the original GRhOOT ontology by~\citet{kuhn2022grhoot}. Classes are shown in blue boxes, individuals in purple boxes, and the blue arrows represent relations. The adapted, reified version of an epiphora is shown in Fig.~\ref{fig:epiphora_new_grhoot}. Individuals are now modeled as classes, while definitions are specific instances. In addition, we simplified the long relations with compound semantics to make them more explicit.

\end{document}